\pdfoutput=1

\documentclass[11pt]{article}

\usepackage[preprint]{acl}

\usepackage{times}
\usepackage{latexsym}

\usepackage[T1]{fontenc}

\usepackage[utf8]{inputenc}

\usepackage{microtype}

\usepackage{inconsolata}

\usepackage{graphicx}

\usepackage{longtable}
\usepackage{CJKutf8}

\title{DeepSeek's WEIRD Behavior: The cultural alignment of Large Language Models and the effects of prompt language and cultural prompting}

\author{James Luther \and Donald Brown \\
  School of Data Science \\
  University of Virginia \\
  Charlottesville, Virginia \\
  \texttt{jluther@virginia.edu} \\\ 
  }

\begin{document}
\maketitle
\begin{abstract}
Culture is a core component of human-to-human interaction and plays a vital role in how we perceive and interact with others. Advancements in the effectiveness of Large Language Models (LLMs) in generating human-sounding text have greatly increased the amount of human-to-computer interaction. As this field grows, the cultural alignment of these human-like agents becomes an important field of study. Our work uses Hofstede's VSM13 international surveys to understand the cultural alignment of the following models: DeepSeek-V3, V3.1, GPT-4, GPT-4.1, GPT-4o, and GPT-5. We use a combination of prompt language and cultural prompting, a strategy that uses a system prompt to shift a model's alignment to reflect a specific country, to align these LLMs with the United States and China. Our results show that DeepSeek-V3, V3.1, and OpenAI's GPT-5 exhibit a close alignment with the survey responses of the United States and do not achieve a strong or soft alignment with China, even when using cultural prompts or changing the prompt language. We also find that GPT-4 exhibits an alignment closer to China when prompted in English, but cultural prompting is effective in shifting this alignment closer to the United States. Other low-cost models, GPT-4o and GPT-4.1, respond to the prompt language used (i.e., English or Simplified Chinese) and cultural prompting strategies to create acceptable alignments with both the United States and China.
\end{abstract}

\section{Introduction}
Culture is a fundamental part of human behavior and provides a shared understanding of how people perceive and interact with the world around them \citep{hofstede_cultures_2024, schein_culture_1991}. Culture affects human priorities, how events are considered in relation to their contextual situation, and how responses affect one's perception in future interactions \citep{oyserman_does_2008}. 

Although perception and reasoning have differences in most cultures, the largest difference can be seen when comparing Western and Eastern societies. These two subsets vary widely in their response to correspondence bias \citep{choi_1999, gilbert_correspondence_1995}, the perception of relationships \citep{ji_culture_2000, peng_nisbett_1999}, and the resolution of conflicting ideas. In the latter context, Eastern cultures support a compromise approach while Western cultures polarize contradictory ideologies to determine the correct response \citep{peng_nisbett_1999}. 

These values and cultural differences were built over time as learned experience and shared understanding passed from generation to generation through the use of language \citep{lotem_evolution_2017}. The language used plays a role in the development and perpetuation of human culture, as it affects the weak cognitive biases that drive many of our perspectives, reasoning, and actions \citep{thompson_culture_2016}. This persists through periods of economic development and technological advancement and plays a role in how these advances are achieved \citep{gelman_how_2017, inglehart_pdf_2024, guiso_does_2006}.

How human-to-human communication is produced digitally, alongside new human-to-computer interaction, has changed drastically via the rise of new artificial intelligence (AI) tools, such as generative auto-reply options, real-time grammar suggestions, and human-sounding language generation tools such as OpenAI's GPT-5 and DeepSeek's V3.1 \citep{hohenstein_artificial_2023, openai_gpt-4_2024, deepseek-ai_deepseek-v3_2025}. Although these tools increase communication efficiency and help to improve stylistic clarity, they can also convey negative connotations to the receiver of this artificially augmented language \citep{hohenstein_artificial_2023} and can be used to create text without human input. 

LLMs such as GPT-5, DeepSeek-V3.1, Claude Opus 4.1, and Gemini have grown in popularity in recent years, and are used in many aspects of life to automate digital communications, in uses ranging from digital personal assistants, chat support clients, automated business communications, news articles, and other forms of online content. It is important to understand the cultural alignment and biases inherent in their training data, as LLMs tend to perpetuate the biases in the data on which they are trained \citep{demszky_using_2023}. Until now, LLMs have shown a bias toward Western, educated, industrialized, rich, and democratic (WEIRD) societies \citep{atari_which_2023, cao_assessing_2023, tao_cultural_2024, wang-etal-2024-countries}, and have had trouble adapting to different cultures, such as Arab nations \citep{masoud_cultural_2025} and Eastern cultures \citep{tao_cultural_2024}. This can cause miscommunications and poor advice when trying to facilitate conflict resolution or collaboration for humans in non-Western cultural settings.

Three different mitigation strategies have been used to shift this alignment towards other cultures. The first and most expensive method is fine-tuning the models to align with a desired culture. Attempts have been made in both Sweden and Japan with limited results \citep{ekgren_gpt-sw3_2024, hornyak_why_2023}. The other two methods, which are the focus of this work, involve the use of prompt language and cultural prompting to alter this alignment. These two were chosen because they have shown promise in previous alignment research \citep{masoud_cultural_2025, zhong_cultural_2024, tao_cultural_2024, kwok_2024} and are an approachable option for changing the alignment of LLM calls without requiring significant resources.

The use of prompt languages other than English has shown success when attempting to alter the cultural alignment of language models developed in Western societies \citep{masoud_cultural_2025, zhong_cultural_2024}. However, the methods used in these experiments were not performed using Hofstede's minimum requirements to calculate his dimensions, specifically a population size of 20 and a minimum number of 10 countries surveyed \citep{hofstede_values_2013}. 

The use of cultural prompts, the act of using system prompts to reflect the desired country of origin, has also shown promise and can be a simple method to alter the default response from a Western-created language model \citep{tao_cultural_2024, kwok_2024}. These results are encouraging, but limitations in population size and the number of models tested limit the impact of these findings as well.

As LLM development has grown and low-cost, state-of-the-art models are improving in both Western and Eastern cultures, it is important to understand the limitations of these mitigation strategies before they are adopted in third-party applications. As interactions with other cultures can influence how an individual interacts with the world \citep{korn_cultural_2014}, knowing the effect of these strategies can aid culturally responsible communication as these systems grow and evolve.

\begin{table*}
  \centering
  \begin{tabular}{ll}
    \hline
    \textbf{Dimension} & \textbf{Equation} \\
    \hline
    \verb|Power Distance Index (PDI)|     & { \(35(m07-m02) + 25(m20-m23) + C_{PDI}\) }           \\
    \verb|Individuality (IDV)|     & { \(35(m04-m01) + 35(m09-m06) + C_{IDV}\) }           \\
    \verb|Masculinity (MAS)|     & { \(35(m05-m03) + 35(m08-m10) + C_{MAS}\) }           \\
    \verb|Uncertainty Avoidance Index (UAI)|     & { \(40(m18-m15) + 25(m21-m24) + C_{UAI}\) }           \\
    \verb|Long-Term Orientation (LTO)|      & { \(40(m13-m14) + 25(m19-m22) + C_{LTO}\) }            \\
    \verb|Indulgence vs. Restraint (IVR)|     & { \(35(m12-m11) + 40(m17-m16) + C_{IVR}\) }           \\\hline
  \end{tabular}
  
  \caption{The equation used to calculate each Hofstede dimension, per Hofstede's VSM 2013 Manual \citep{hofstede_values_2013}. \(m01\) indicates the mean value for all answers to Question 01 in a given population. The constants are used to normalize the range of the dimension values to 0-100; they can be found in the Appendix.}
  \label{tab:accents}
\end{table*}

\section{Methods}
In this work, we use Hofstede's VSM13 International Survey and its results \citep{hofstede_cultures_2024} to prompt and measure the cultural alignment of LLMs against the United States and China. Using these surveys, we prompt six prominent models, GPT-4 \citep{openai_gpt-4_2024}, GPT-4.1, GPT-4o \citep{openai_gpt-4o_2024}, GPT-5, DeepSeek-V3 \citep{deepseek-ai_deepseek-v3_2025}, and DeepSeek-V3.1. The VSM13 survey questions were slightly modified to ensure a response from the models by shifting the subject to the average person, not the survey respondent themselves ('Do you feel' becomes 'Does the average person feel'). The survey questions and system prompts, in both English and Simplified Chinese, can be found in the Appendix.

Hofstede's methods for creating his cultural dimensions, as specified in the VSM13 Manual \citep{hofstede_values_2013}, set the minimum population size as 20. We adopted this number, while noting that others using his dimensions fell below this recommended threshold \citep{masoud_cultural_2025, zhong_cultural_2024}.

We treated each of these models as six distinct populations and prompted them using the following format: in English without cultural prompting (model\_en), in English with US cultural prompting (model\_en\_US), in English with Chinese cultural prompting (model\_en\_CH), in Simplified Chinese with no cultural prompting (model\_sc), in Simplified Chinese with US cultural prompting (model\_sc\_US), and in Simplified Chinese with Chinese cultural prompting (model\_sc\_CH). Because these six distinct methods were treated as standalone populations, each required 20 complete survey responses. 

We prompted these languages via their respective APIs in batches of 5 surveys at a time, with each survey containing 24 questions, for a total of 120 questions per batch. Each batch was run four separate times, totaling a population of 20 survey responses per set of model-language-cultural prompts. For each model, this totaled 2,880 prompts and responses. As there are six models used in this work, 17,280 prompts and responses were recorded and analyzed in total. The temperature hyperparameter for each model was set to 2, as variation in the models' responses is key to simulating the breadth one expects from the general population of any given country.

\begin{figure*}[t]
  \includegraphics[width=0.48\linewidth]{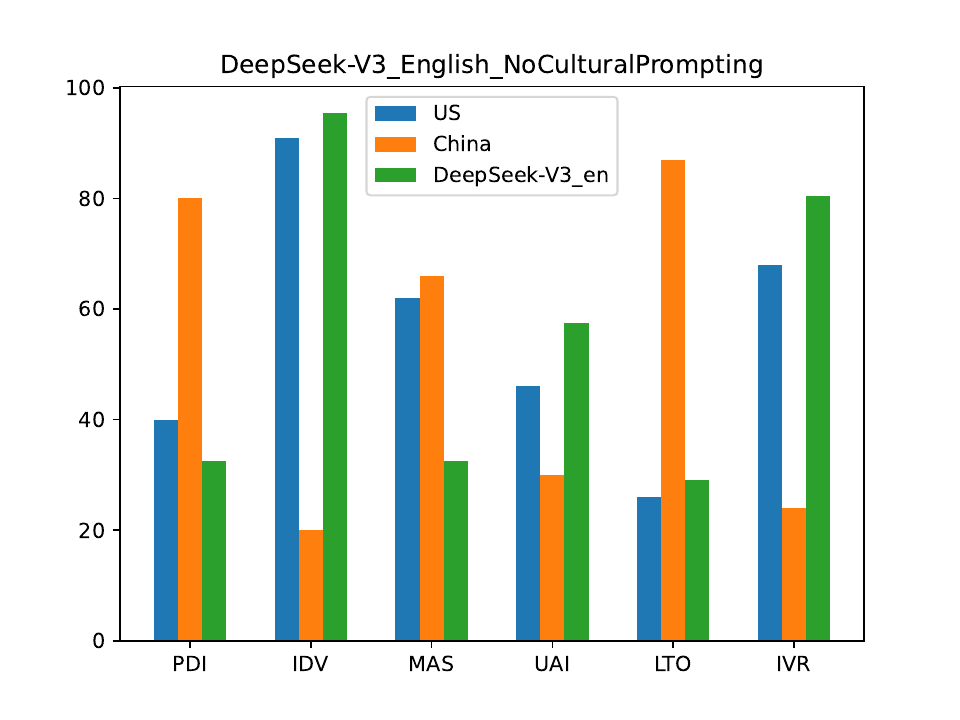} \hfill
  \includegraphics[width=0.48\linewidth]{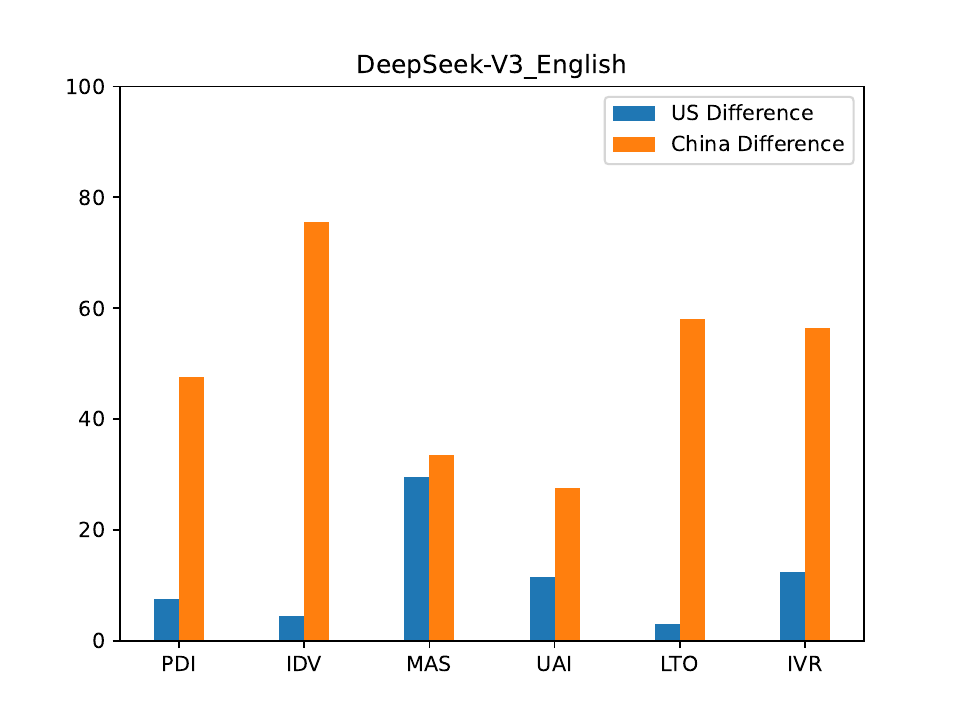}
  \caption {The DeepSeek-V3 model's calculated dimensions alongside US and China's Hofstede dimensions (left) and the difference between the model's responses and each corresponding country (right).}
\end{figure*}

\begin{figure*}[t]
  \includegraphics[width=0.48\linewidth]{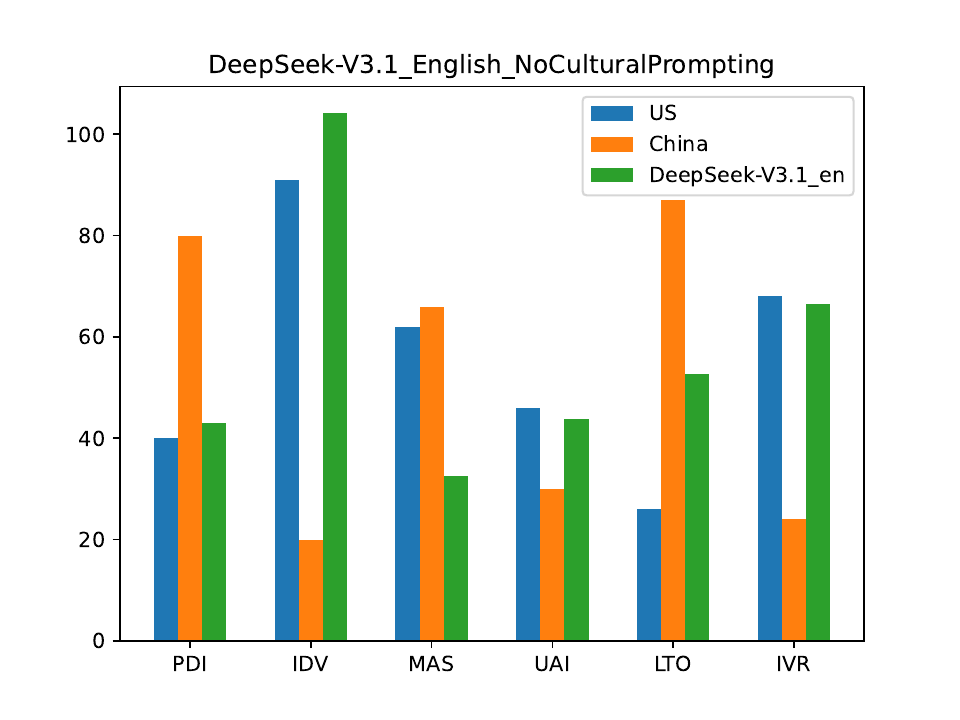} \hfill
  \includegraphics[width=0.48\linewidth]{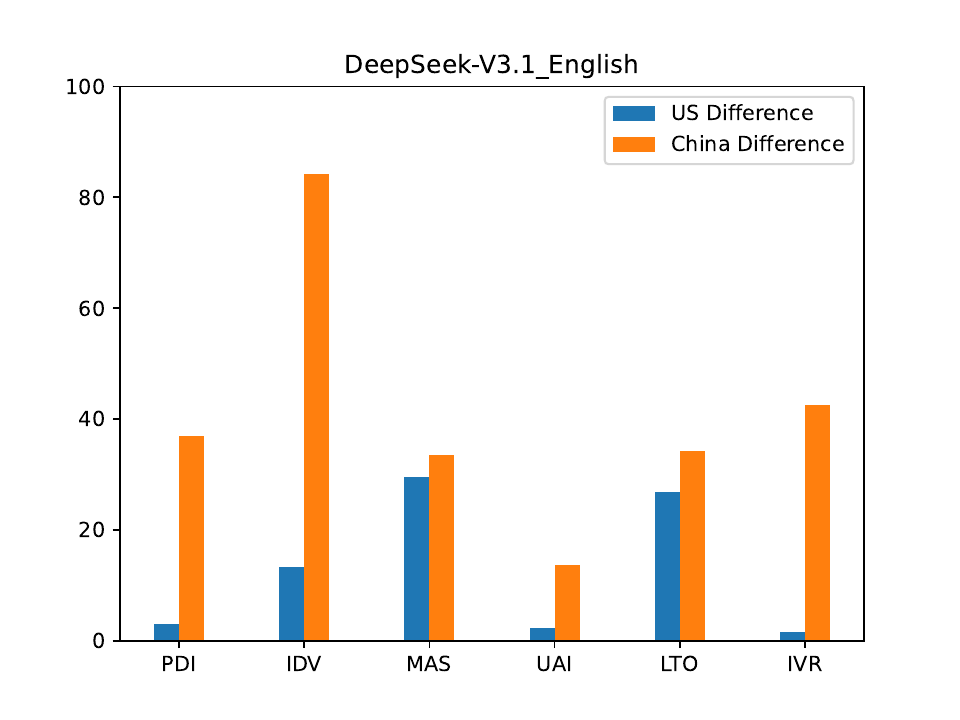} 
  \caption {The DeepSeek-V3.1 model's calculated dimensions alongside US and China's Hofstede dimensions (left) and the difference between the model's responses and each corresponding country (right).}
\end{figure*}


The results of each question within a population were averaged to a mean via Hofstede's instructions \citep{hofstede_values_2013}, and the dimensions were generated using the equations found in Table 1. The constants used are designed to move the range of all calculated results to a minimum of 0 and a maximum of 100. When there was a range smaller or larger than 100, the selected constant pushes the midpoint of this range to 50. These constants can be found in the Appendix.

The metric used for measuring a model's alignment is the sum of the absolute values of the distance from each dimension value to the corresponding country's dimensions. 


\begin{table*}
  \centering
  \begin{tabular}{| c | cc | cc | c |}
    \hline
    \textbf{Country} & \textbf{Category} & \textbf{Total} & \textbf{Category} & \textbf{Total} & \textbf{Improvement} \\
    \hline
    {US} & {Simp. Chinese} & {867.75} & {English} & {\textbf{654.75}} & {24.6\%}  \\
    {US} & {English} & {\textbf{654.75}} & {+ US Prompting} & {682.25} & {-4.2\%}  \\
    {US} & {Simp. Chinese} & {867.75} & {+ US Prompting} & {855} & {1.5\%}  \\
    {China} & {English} & {1349.75} & {Simp. Chinese} & {1145.25} & {15.2\%}  \\
    {China} & {Simp. Chinese} & {1145.25} & {+ Chinese Prompting} & {1059} & {7.5\%}  \\
    {China} & {English} & {1349.75} & {+ Chinese Prompting} & {\textbf{947.75}} & {29.8\%}  \\
    \hline
  \end{tabular}
  
  \caption{The sum distance of all models for a given prompting style to the measured Hofstede dimensions of the specified country (US or China), with the measured improvement shown as a percentage. The first column corresponds to the specified prompt language without cultural prompting, while the second shifts to a different prompt language or adds the given country's cultural prompting to the same prompt language. The bold value represents the lowest total found across all cultural adaptation techniques for the US and China. }
  \label{tab:accents}
\end{table*}

\section{Results}
Across the six models tested, most were able to create a strong alignment with the United States, as measured by having an average dimension distance of 15 or less across the six dimensions from Hofstede's VSM13 survey. When prompted in English without cultural prompting, four of the six models achieved a strong alignment, including DeepSeek-V3 (68.5), DeepSeek-V3.1 (76.25), GPT-5 (76.75), and GPT-4o (86.5). A soft alignment, or an average dimension distance of 15 to 20, can be found with nine other populations of varying languages and cultural prompting strategies. GPT-4 was the only model unable to align with the United States. The full table of results can be found in the Appendix.

Only one model was able to align closely with China, GPT-4.1 (89.25), when prompted in Simplified Chinese. In addition, there were only two soft alignments found, with GPT-4o (English with cultural prompting) and GPT-4.1 (Simplified Chinese with cultural prompting). The overall distance found to China was larger across all models and reflects a bias towards the United States, even for models created by the Chinese company DeepSeek. 

Using the specified cultural shifting methods, the alignment improvement strategies were largely successful. The results for each method can be found in Table 2. The shift from Simplified Chinese to native English improved the overall distance to the United States across all models by 24.6\%, while the same English to native Simplified Chinese improved alignment with China by 15.2\%. In addition to the benefit of using the country's native language, adding cultural prompting improved the alignment with China by an additional 7.5\%. When adding cultural prompting to native English and measuring against the dimensions of the US, we see a slightly negative result. This is due to the models' strong alignment in English without cultural prompting, which left little room for improvement. Of the two models that did not have a strong alignment with the United States, both showed an average improvement of 33.6\%. 

When using cultural prompting and a non-native language, the improvement depends on the language. When using English and culturally prompting for China, the alignment improved by 29.8\% compared to English without cultural prompting. When using Simplified Chinese and prompting for the US, the alignment improved by only 1.5\% over Simplified Chinese with no cultural prompting. The models show significantly more flexibility when prompted in English, leading to larger gains from cultural prompting strategies when compared to the same techniques using Simplified Chinese.

While the results for all models weighed together were positive, the results by model varied widely. DeepSeek-V3 showed an alignment with the dimension data of the United States, with five of the six methods resulting in a strong or soft alignment with the US, and all six aligning closer to the US than China. The dimension data, as shown in Figure 1, clearly show a strong alignment with the US dimension data from Hofstede \citep{hofstede_cultures_2024}. In addition, each of the methods used to shift this alignment toward China failed to achieve a closer alignment to China than the United States. This result is reinforced by the responses from DeepSeek-V3.1, which showed a similarly close alignment with the United States. Four of the six methods tested resulted in a strong or soft alignment with the US, while all six again resulted in a closer alignment to the US than China. DeepSeek's unique training method \citep{deepseek-ai_deepseek-v3_2025} appears to limit its ability to shift its cultural alignment through the selected methods, and its models align very closely with the United States. All distances can be found in the Appendix.

GPT-4 showed no strong or soft alignments with either the United States or China using any method tested. When prompted in English without cultural prompting, it surprisingly favored China (127.75 vs. 191.75), though not enough to consider it a strong or soft alignment. When culturally prompting for the United States, the alignment shifted to a closer alignment with the US (124.75 vs. 189.25). The relative distance in total was larger compared to other models, showing that GPT-4 remains largely neutral and does not shift its alignment strongly to either country. GPT-5, on the other hand, shows a strong alignment with the United States when prompted in English with no cultural prompting (76.75), alongside a soft US alignment when prompted in Simplified Chinese (98.5). When culturally prompted for China, the distance was shortened, but not enough to approach a soft alignment. 

GPT-4.1 showed a strong alignment with China when prompted in Simplified Chinese (89.25), with English favoring the US but not significantly. This model's strength came when prompted in the native language of the country alongside cultural prompting, with both the US and China showing a soft alignment using this method. GPT-4o, when prompted in English, produced a strong alignment with the United States (86.5), while prompts in Simplified Chinese bridged the distance marginally but favored the US. When prompting in English with cultural prompting to the desired country, the alignment responded accordingly and showed a soft alignment for both the US (93) and China (114.75).

In general, cultural prompting was much more successful in English than it was in Simplified Chinese. When looking at the full effect on both the US and China and controlling for language, we find that the English-language cultural prompts were able to achieve an 18.7\% decrease in total alignment distance. When focusing on the Simplified Chinese cultural prompts, we only see an improvement of 4.9\%. This is due to the significant improvement when English prompts were culturally prompted for China, a non-native speaking country. This method was highly effective at shifting the cultural alignment of models' responses, and is easy to integrate into existing applications that communicate with individuals from non-native English cultures.

\section{Conclusion}
Overall, all models favored the United States and had much lower total distances throughout. In each model tested, at least one of the six populations (model, prompt language, and cultural prompt) had a total distance to China of greater than 200. Across all populations, the total distance to the US only exceeded 200 once. This general bias towards the United States enforces findings from \citeauthor{atari_which_2023} and works to quantify that bias in different models. Models from the Chinese company DeepSeek showing this same bias is a significant finding, reflecting that the source of this bias is not limited to where the model is created. 

These results highlight the malleability of some models, like GPT-4o and GPT-4.1, to adapt to China's culture, as well as the inability of others, such as GPT-4, GPT-5, and both DeepSeek models. As low-cost models such as GPT-4o, GPT-4.1, and DeepSeek-V3.1 are integrated into more third-party applications and utilized in everyday life, the alignment of one's culture with the model with which they are interacting becomes more important. US-based models GPT-4o and GPT-4.1 showed the ability to adapt successfully using either method, which can lead to low-cost implementation of culturally-responsive models in third-party applications.

GPT-4's inability to achieve a strong or soft US alignment is significant, as well as its closer alignment with China when prompted in English with no cultural prompting. While its alignment with China was better than expected, it was the only model with zero alignment distances that fell under 120 to either country. Its poor performance in shifting its alignment using low-cost cultural prompting shows it as an ineffective model for aligning with the US or China, despite the complexity of the model and the depth of its training \citep{openai_gpt-4_2024}. 

\section{Limitations}
Hofstede's VSM13 instructions provide two key minimum requirements to complete this analysis: population size (20) and number of populations included (10). We completed this work with a population size of the minimum 20. While this meets the minimum requirement as specified by Hofstede, this is far lower than the number of surveys collected from each participating country in the VSM13 International Survey. The inclusion of more surveys could provide more accurate results, as the full breadth of responses could be viewed, and a larger sample size could lead to more accurate mean question values, which are used in the creation of each dimension.

In addition, this work was completed using only two countries for reference (the United States and China), two languages (English and Simplified Chinese), and six models (GPT-4, GPT-4.1, GPT-4o, GPT-5, DeepSeek-V3, and DeepSeek-V3.1). This limited scope shows how these models align with their originating countries, but could provide closer alignment distances to other native-speaking countries, such as the United Kingdom, Australia, or Singapore. We could also expand the number of languages used to prompt the language models with other popular languages, such as Spanish, Hindi, and Japanese. Many more language models can be tested using this framework, including popular choices from Gemini, Llama, and Claude.

\bibliography{custom}

\appendix

\section{Appendix}
\label{sec:appendix}

\begin{table*}
  \centering
  \begin{tabular}{l|cc|cccccc}
    \hline
    \textbf{Model} & \textbf{US Distance} & \textbf{China Distance} & \textbf{PDI} & \textbf{IDV} & \textbf{MAS} & \textbf{UAI} & \textbf{LTO} & \textbf{IVR}\\
    \hline
    {GPT-5\_sc\_CH} & {182.25} & {148.75} & {44.5} & {48.25} & {48.25} & {71.25} & {95.5} & {41.5} \\
    {GPT-5\_sc\_US} & {147.5} & {285.5} & {7.25} & {99.0} & {43.0} & {31.25} & {60.25} & {106.75} \\
    {GPT-5\_sc} & {98.5} & {213.0} & {10.75} & {81.5} & {53.5} & {30.0} & {52.0} & {58.75} \\
    {GPT-5\_en\_CH} & {268.25} & {164.25} & {85.5} & {-6.0} & {57.0} & {75.0} & {41.0} & {-8.75} \\
    {GPT-5\_en\_US} & {120.75} & {276.25} & {5.75} & {86.75} & {64.0} & {23.75} & {44.0} & {108.0} \\
    {GPT-5\_en} & {76.75} & {222.75} & {22.5} & {85.0} & {65.75} & {26.25} & {45.25} & {78.5} \\
    {DSV3.1\_sc\_CH} & {102.0} & {221.5} & {25.5} & {58.75} & {64.0} & {71.25} & {50.0} & {72.0} \\
    {DSV3.1\_sc\_US} & {117.25} & {250.75} & {25.0} & {90.25} & {27.25} & {43.0} & {73.5} & {84.25} \\
    {DSV3.1\_sc} & {145.25} & {195.75} & {30.25} & {64.0} & {39.5} & {66.25} & {79.5} & {55.75} \\
    {DSV3.1\_en\_CH} & {155.0} & {170.5} & {34.5} & {43.0} & {85.0} & {66.25} & {51.25} & {35.0} \\
    {DSV3.1\_en\_US} & {119.5} & {317.5} & {13.25} & {106.0} & {15.0} & {28.75} & {29.0} & {78.5} \\
    {DSV3.1\_en} & {76.25} & {245.25} & {43.0} & {104.25} & {32.5} & {43.75} & {52.75} & {66.5} \\
    {GPT-4\_sc\_CH} & {167.5} & {230.0} & {15.75} & {46.5} & {48.25} & {88.75} & {44.0} & {43.75} \\
    {GPT-4\_sc\_US} & {152.75} & {251.25} & {20.5} & {46.5} & {44.75} & {82.5} & {50.25} & {78.75} \\
    {GPT-4\_sc} & {185.25} & {227.75} & {0.0} & {43.0} & {58.75} & {91.25} & {72.75} & {66.0} \\
    {GPT-4\_en\_CH} & {165.0} & {161.0} & {48.25} & {34.25} & {74.5} & {82.5} & {44.0} & {35.0} \\
    {GPT-4\_en\_US} & {124.75} & {189.25} & {48.5} & {53.5} & {37.75} & {58.75} & {65.75} & {70.0} \\
    {GPT-4\_en} & {191.75} & {127.75} & {57.5} & {51.75} & {39.5} & {56.25} & {96.0} & {35.75} \\
    {GPT-4.1\_sc\_CH} & {175.75} & {93.25} & {96.5} & {44.75} & {65.75} & {32.5} & {45.25} & {31.5} \\
    {GPT-4.1\_sc\_US} & {186.75} & {173.75} & {94.25} & {76.25} & {44.75} & {31.25} & {81.0} & {98.75} \\
    {GPT-4.1\_sc} & {182.75} & {89.25} & {93.0} & {51.75} & {71.0} & {32.5} & {71.75} & {45.75} \\
    {GPT-4.1\_en\_CH} & {178.5} & {160.5} & {100.0} & {46.5} & {48.25} & {63.75} & {44.0} & {43.5} \\
    {GPT-4.1\_en\_US} & {105.5} & {259.5} & {17.5} & {81.5} & {32.5} & {32.5} & {44.0} & {80.5} \\
    {GPT-4.1\_en} & {155.0} & {240.0} & {11.25} & {78.0} & {32.5} & {8.75} & {44.0} & {39.5} \\
    {GPT-4o\_sc\_CH} & {136.75} & {122.25} & {78.25} & {62.25} & {53.5} & {33.75} & {72.0} & {71.0} \\
    {GPT-4o\_sc\_US} & {158.5} & {169.0} & {66.0} & {76.25} & {71.0} & {35.75} & {83.75} & {108.75} \\
    {GPT-4o\_sc} & {136.5} & {186.0} & {65.25} & {93.75} & {53.5} & {33.75} & {81.75} & {100.0} \\
    {GPT-4o\_en\_CH} & {129.75} & {114.75} & {82.75} & {43.0} & {60.5} & {39.5} & {42.75} & {53.75} \\
    {GPT-4o\_en\_US} & {93.0} & {196.5} & {38.0} & {51.75} & {39.5} & {30.5} & {37.5} & {70.25} \\
    {GPT-4o\_en} & {86.5} & {215.5} & {36.75} & {60.5} & {43.0} & {40.0} & {43.0} & {78.75} \\
    {DSV3\_sc\_CH} & {63.25} & {243.25} & {50.0} & {97.25} & {57.0} & {57.5} & {44.0} & {80.5} \\
    {DSV3\_sc\_US} & {92.25} & {241.25} & {25.5} & {79.75} & {51.75} & {57.5} & {58.25} & {80.5} \\
    {DSV3\_sc} & {119.5} & {233.5} & {39.5} & {102.5} & {46.5} & {57.5} & {94.0} & {80.5} \\
    {DSV3\_en\_CH} & {139.25} & {176.75} & {28.25} & {9.75} & {67.5} & {57.5} & {21.0} & {43.75} \\
    {DSV3\_en\_US} & {118.75} & {283.25} & {32.5} & {55.25} & {32.5} & {57.5} & {4.0} & {80.5} \\
    {DSV3\_en} & {68.5} & {298.5} & {32.5} & {95.5} & {32.5} & {57.5} & {29.0} & {80.5} \\					
    \hline
  \end{tabular}
  
  \caption{The dimension values for all model-language-culture populations. Prompt language is designated by \_en (English) or \_sc (simplified Chinese). Cultural prompting is designated as \_US (US cultural prompt), \_CH (China cultural prompt), or blank (no cultural prompting).}
  \label{tab:accents}
\end{table*}

\begin{table*}
  \centering
  \begin{tabular}{ll}
    \hline
    \textbf{Constant} & \textbf{Value} \\
    \hline
    {\(C_{PDI}\)}     & {15}           \\
    {\(C_{IDV}\)}     & {11.5}        \\           
    {\(C_{MAS}\)}     & {67.5}           \\           
    {\(C_{UAI}\)}     & {82.5}           \\           
    {\(C_{LTO}\)}     & {44}           \\            
    {\(C_{IVR}\)}     & {45.5}       \\
    \hline
  \end{tabular}
  
  \caption{The constants used to correct the range of each dimension calculation.}
  \label{tab:accents}
\end{table*}

\begin{figure*}[t]
  \includegraphics[width=\linewidth]{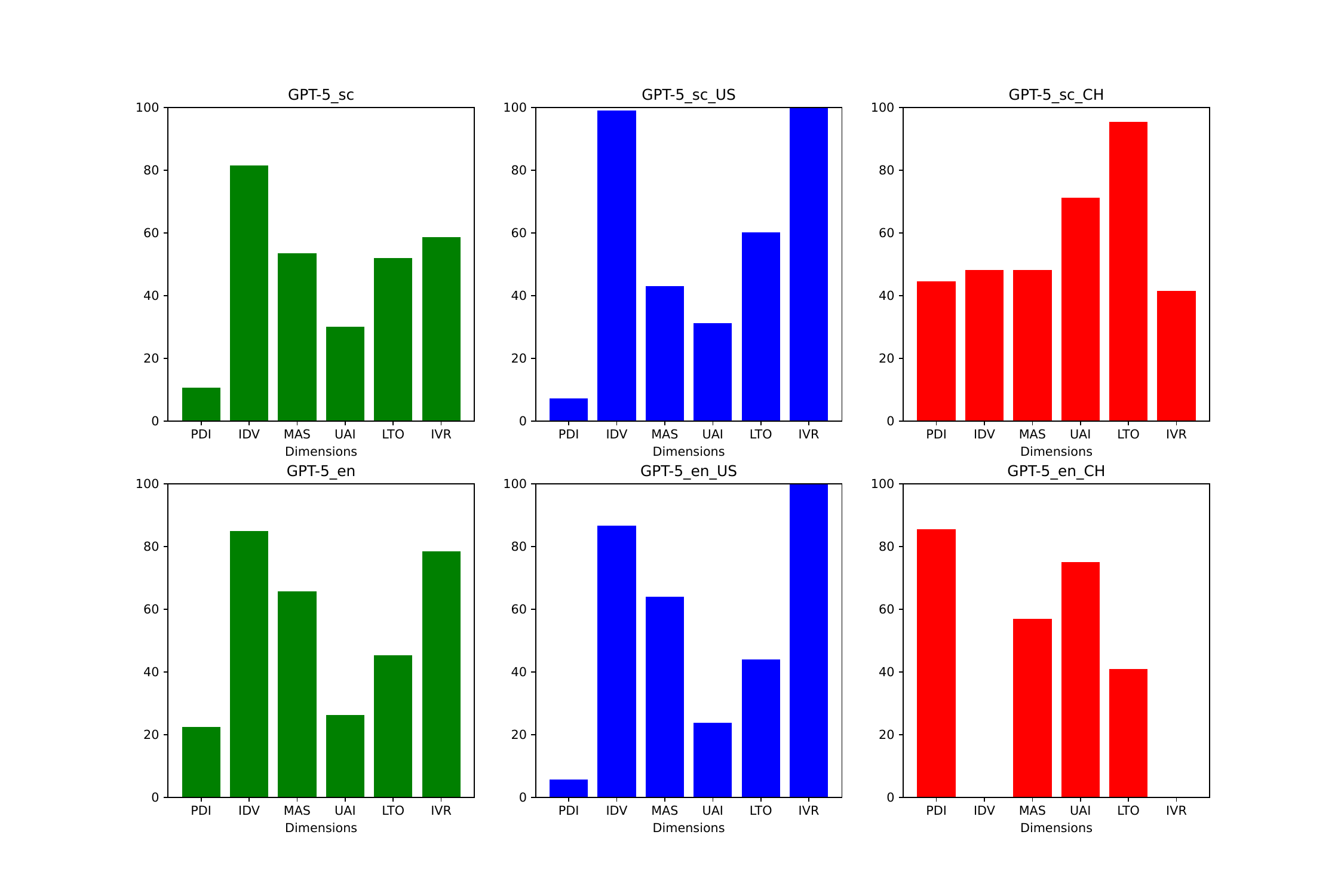} 
  \caption {All calculated GPT-5 dimensions. Lowercase abbreviations indicate prompt language, Uppercase abbreviations indicate cultural prompting method (no cultural prompting used if blank). }
\end{figure*}

\begin{figure*}[t]
  \includegraphics[width=\linewidth]{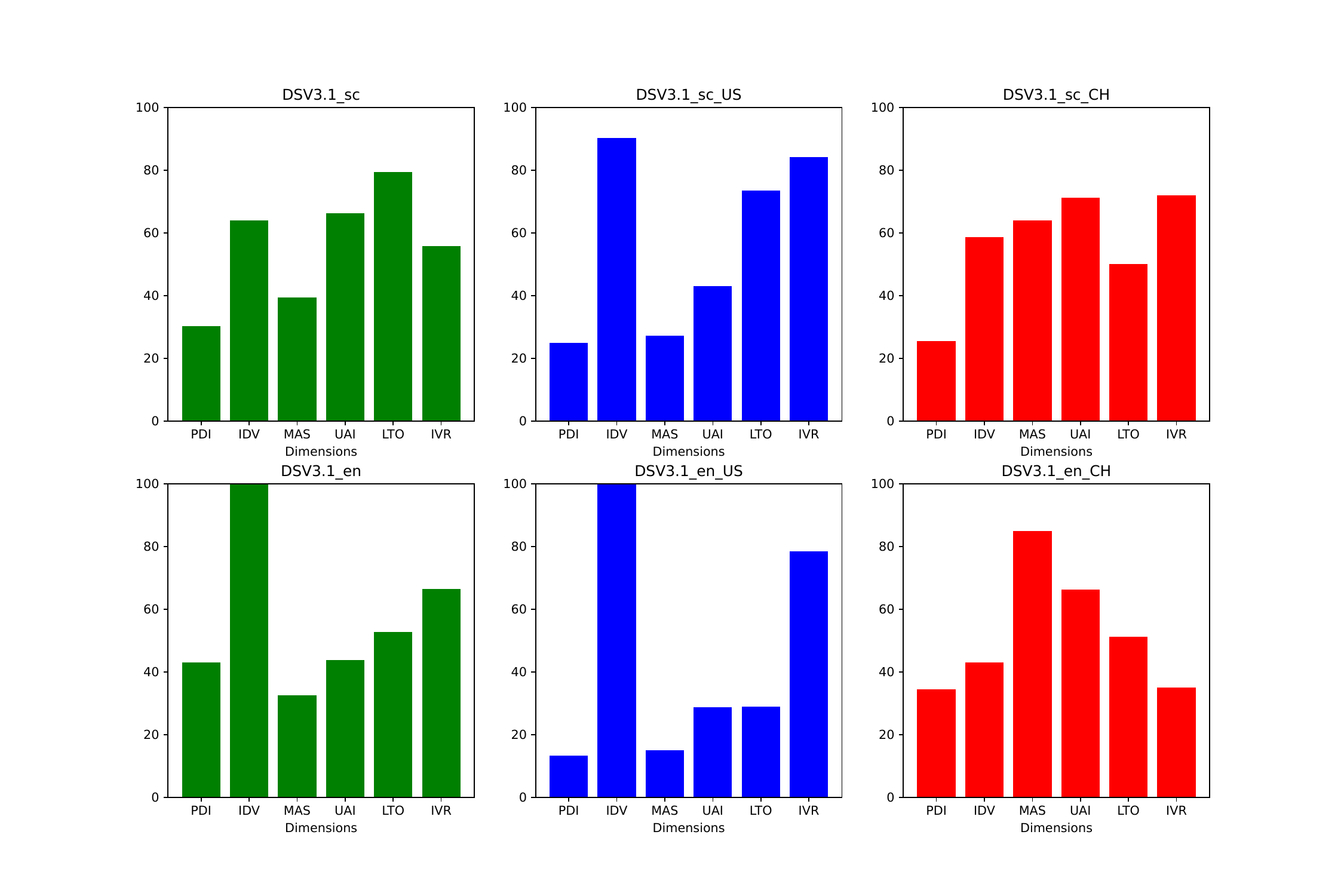} 
  \caption {All calculated DeepSeek-V3.1 dimensions. Lowercase abbreviations indicate prompt language, Uppercase abbreviations indicate cultural prompting method (no cultural prompting used if blank). }
\end{figure*}

\begin{figure*}[t]
  \includegraphics[width=\linewidth]{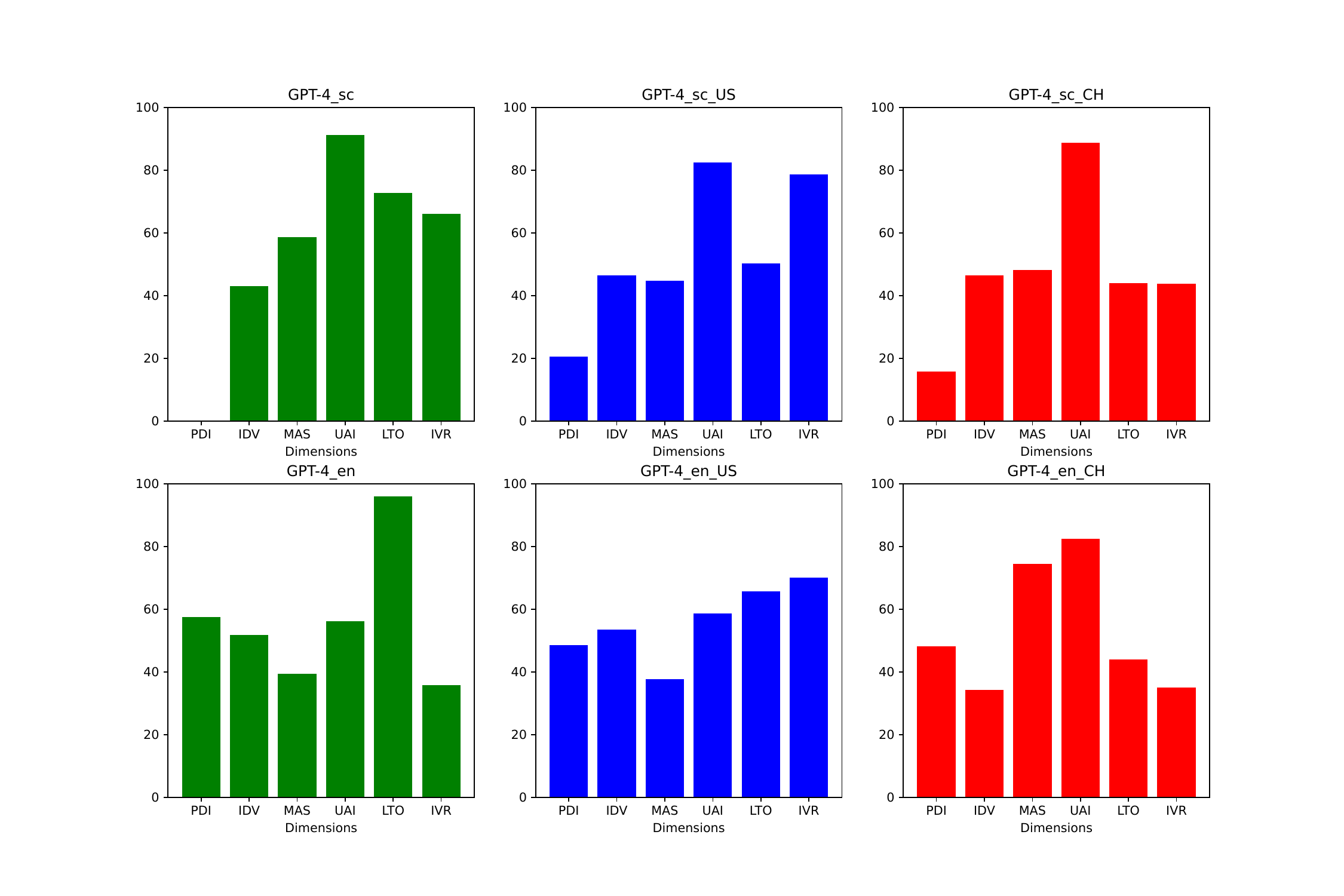} 
  \caption {All calculated GPT-4 dimensions. Lowercase abbreviations indicate prompt language, Uppercase abbreviations indicate cultural prompting method (no cultural prompting used if blank). }
\end{figure*}

\begin{figure*}[t]
  \includegraphics[width=\linewidth]{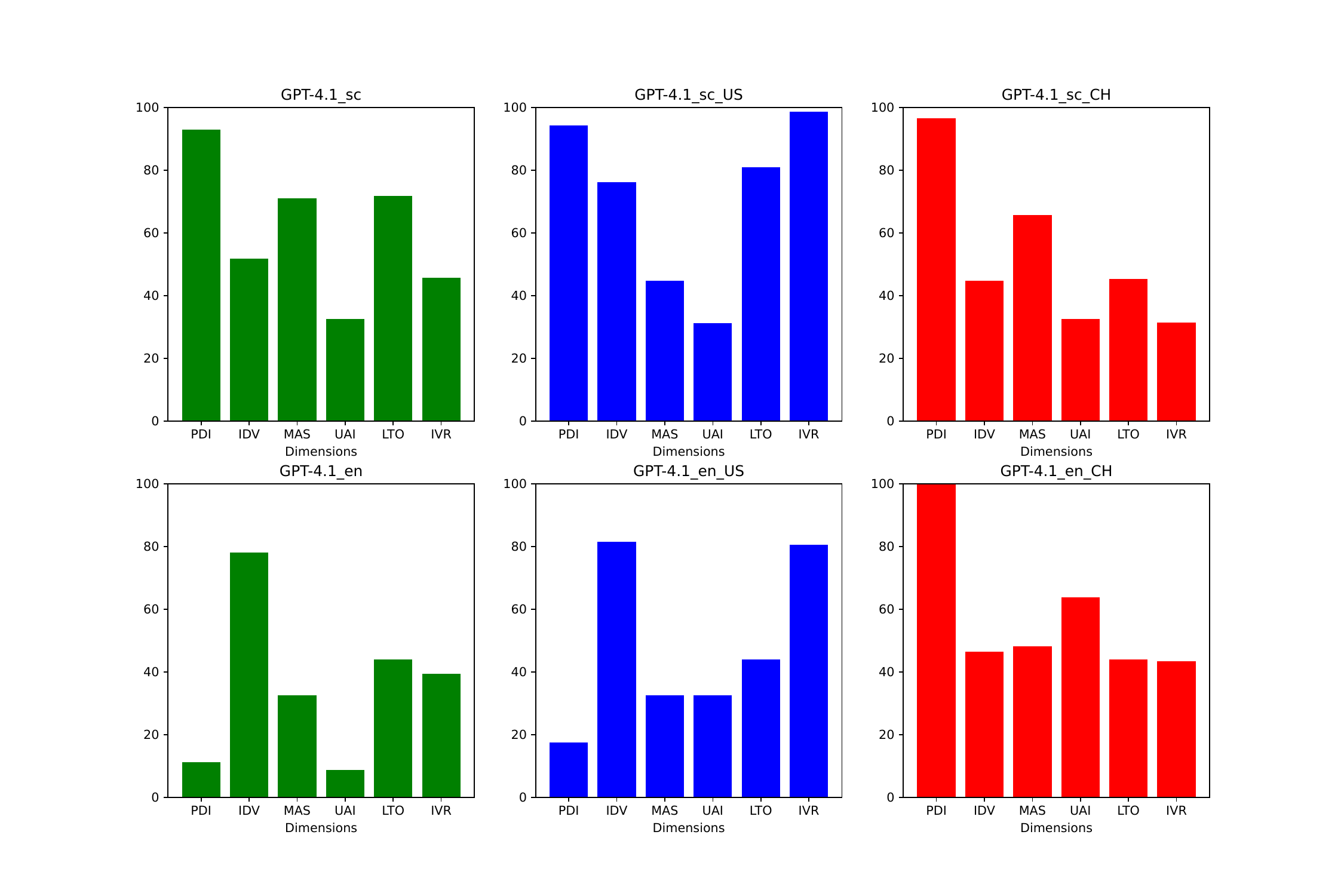} 
  \caption {All calculated GPT-4.1 dimensions. Lowercase abbreviations indicate prompt language, Uppercase abbreviations indicate cultural prompting method (no cultural prompting used if blank). }
\end{figure*}

\begin{figure*}[t]
  \includegraphics[width=\linewidth]{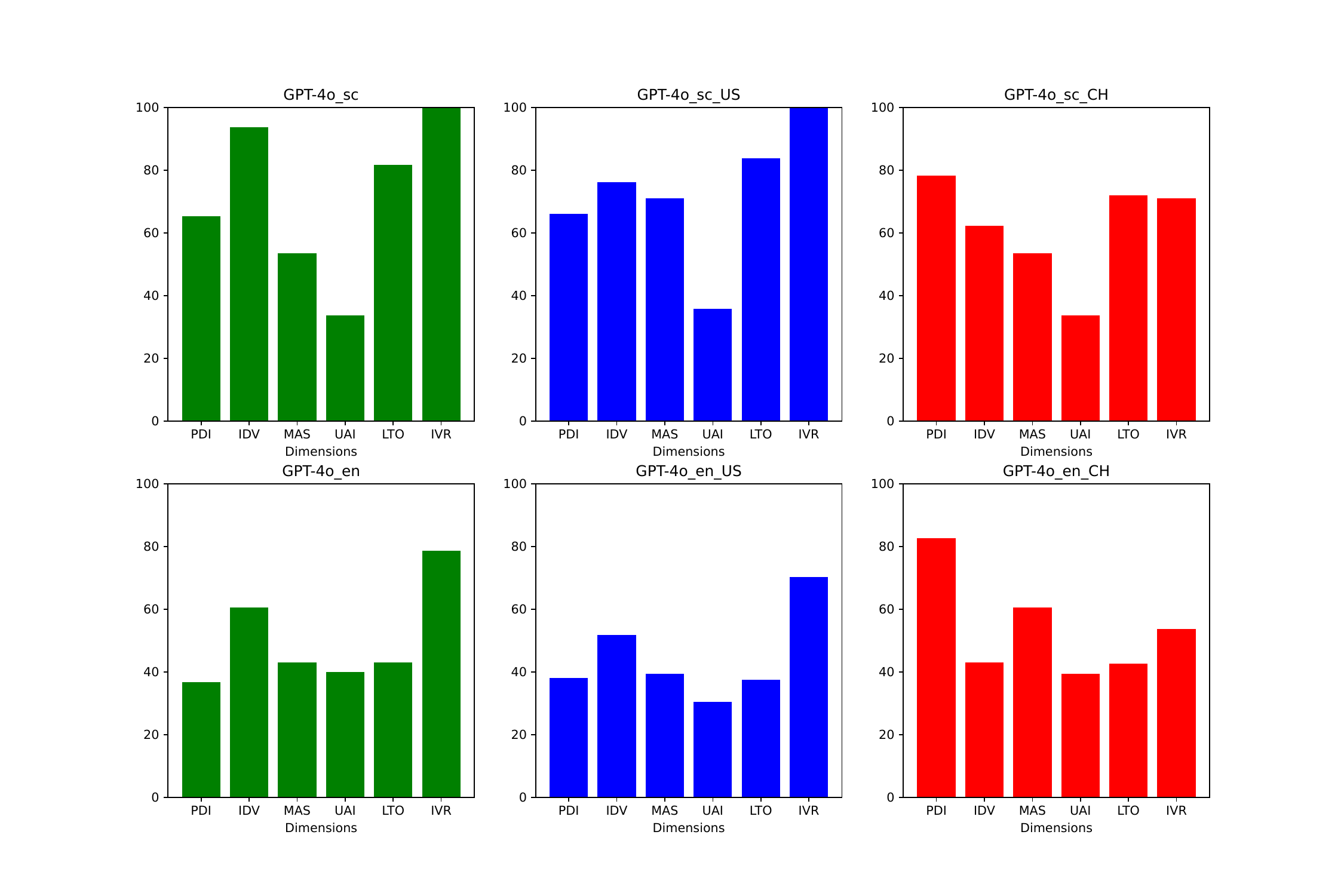} 
  \caption {All calculated GPT-4o dimensions. Lowercase abbreviations indicate prompt language, Uppercase abbreviations indicate cultural prompting method (no cultural prompting used if blank). }
\end{figure*}

\begin{figure*}[t]
  \includegraphics[width=\linewidth]{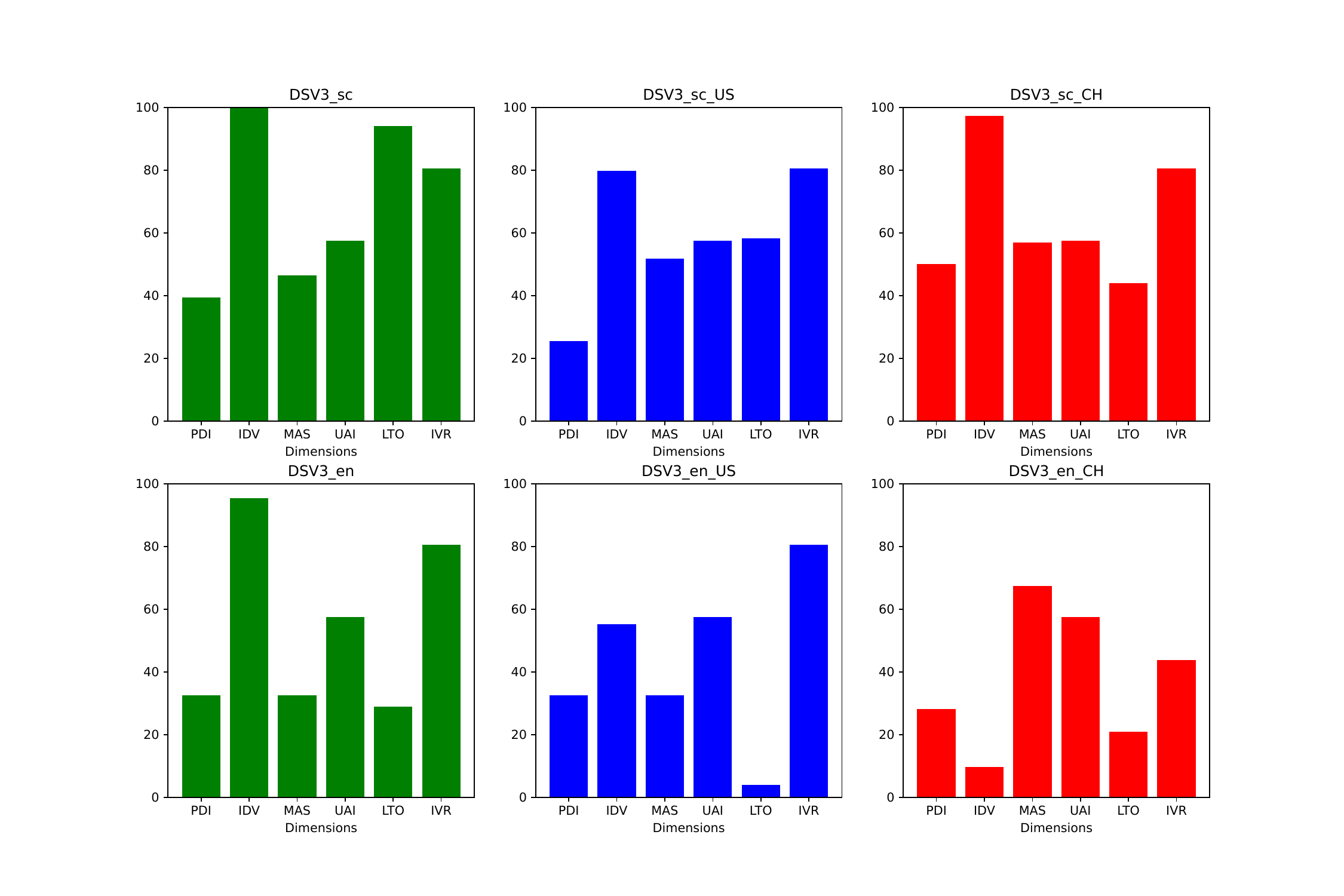} 
  \caption {All calculated DeepSeek-V3 dimensions. Lowercase abbreviations indicate prompt language, Uppercase abbreviations indicate cultural prompting method (no cultural prompting used if blank). }
\end{figure*}

\begin{figure*}[t]
  \includegraphics[width=\linewidth]{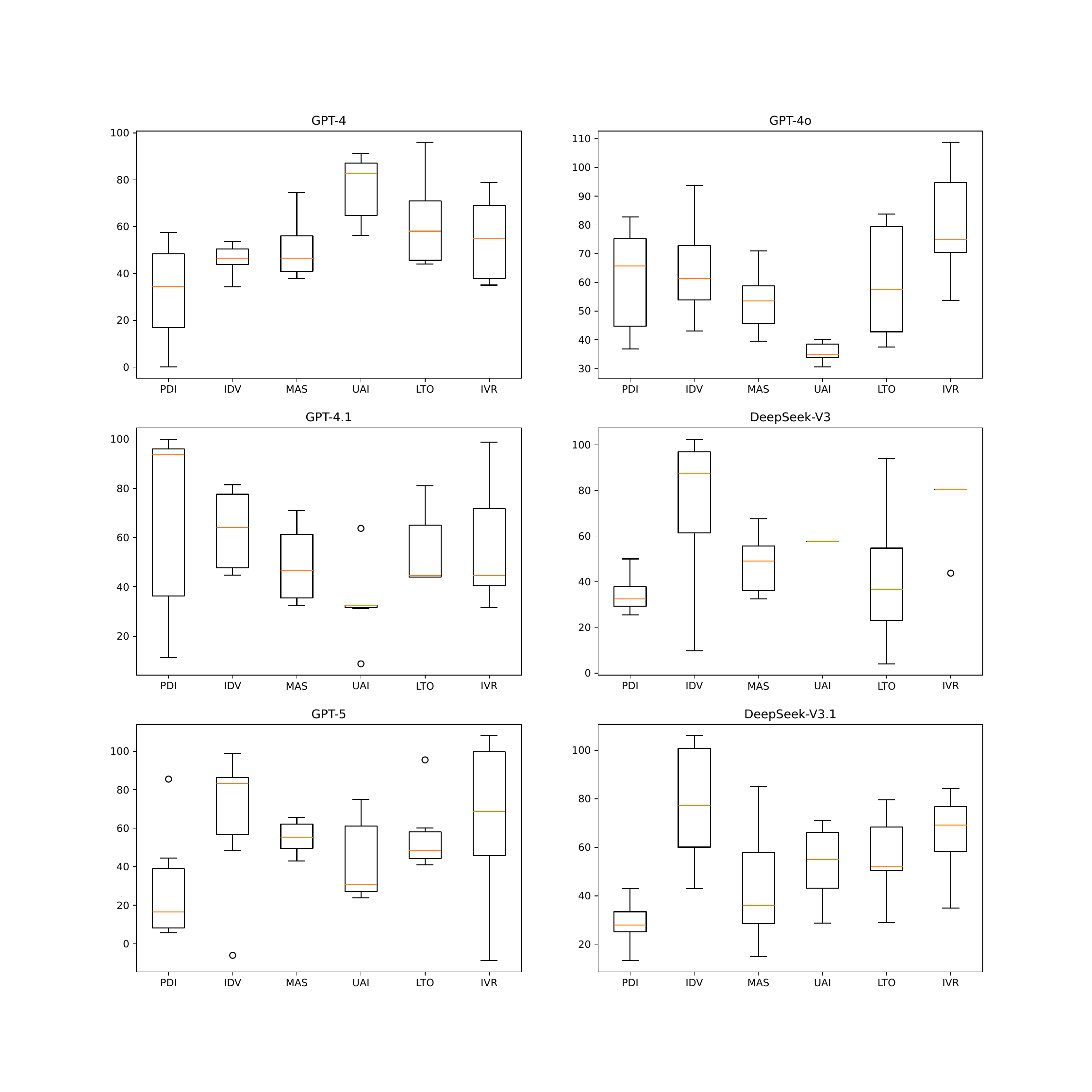} 
  \caption {Box Plot of the calculated dimensions by model. This highlights the variability in each dimension's calculated value based on the selected alignment adjustment methods.}
\end{figure*}

\begin{figure*}[t]
  \includegraphics[width=\linewidth]{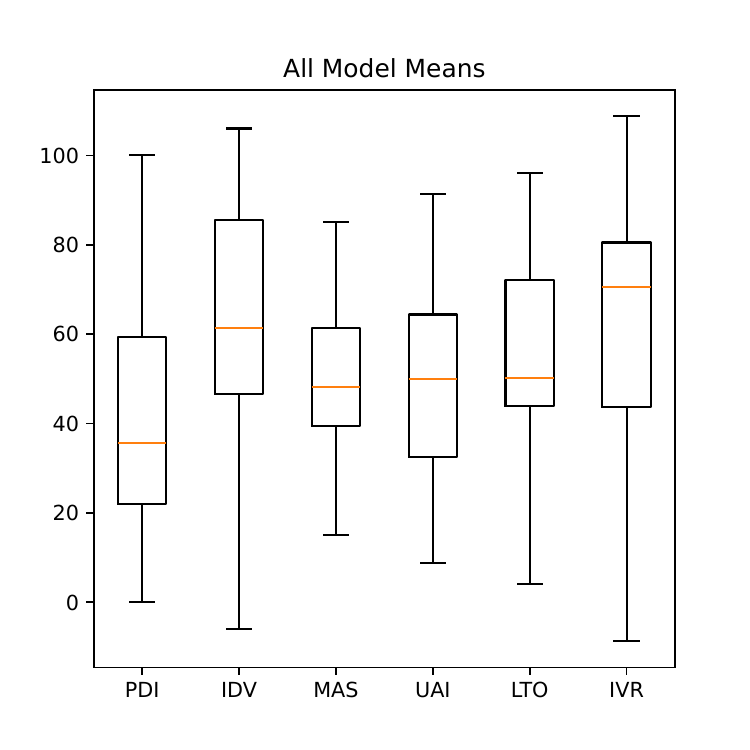} 
  \caption {Box Plot of the calculated dimensions for all models. This shows that the models and their adjusted methods fill the breadth of possible answers for most dimensions.}
\end{figure*}

\begin{table*}
  \centering
  \begin{tabular}{l p{5in}}
    \hline
    \textbf{Culture} & \textbf{System Prompt} \\
    \hline
    {None}     & {You are an average person answering a survey question.}           \\
    {US}     & {You are an average person from the United States answering a survey question.}        \\           
    {China}     & {You are an average person from China answering a survey question.}           \\           
    \hline
  \end{tabular}
  
  \caption{The English language system prompts used for cultural prompting.}
  \label{tab:accents}
\end{table*}

\begin{CJK*}{UTF8}{gbsn}
\begin{table*}
  \centering
  \begin{tabular}{l p{5in}}
    \hline
    \textbf{Culture} & \textbf{System Prompt} \\
    \hline
    {None}     & {您是一位正在回答调查问题的普通人。}           \\
    {US}     & {您是来自美国的普通人，正在回答调查问题。}        \\           
    {China}     & {您是来自中国的一名普通人，正在回答一项调查问题。}           \\           
    \hline
  \end{tabular}
  
  \caption{The simplified Chinese language system prompts used for cultural prompting.}
  \label{tab:accents}
\end{table*}
\end{CJK*}

\clearpage
\onecolumn
\begin{longtable}{c p{5in}}
  \caption{The English language question prompts used to survey the selected models.}\label{tab:label} \\
  \hline
  \textbf{Question} & \textbf{Prompt Text} \\
  \hline
  \endhead
  \hline
  \endfoot  
  {1}     & {Question: Please think of an ideal job, disregarding your present job. In choosing an ideal job, how important would it be to have sufficient time for your personal or home life? Please use a scale from 1 to 5, where 1 is of utmost importance, 2 is very important, 3 is of moderate importance, 4 is of little importance, 5 is of very little or no importance. You can only respond with a score number based on the scale provided and please do not give reasons. Your score number:}       \\
    {2}     & {Question: Please think of an ideal job, disregarding your present job. In choosing an ideal job, how important would it be to have a boss (direct superior) you can respect? Please use a scale from 1 to 5, where 1 is of utmost importance, 2 is very important, 3 is of moderate importance, 4 is of little importance, 5 is of very little or no importance. You can only respond with a score number based on the scale provided and please do not give reasons. Your score number:}       \\           
    {3}     & {Question: Please think of an ideal job, disregarding your present job. In choosing an ideal job, how important would it be to get recognition for good performance? Please use a scale from 1 to 5, where 1 is of utmost importance, 2 is very important, 3 is of moderate importance, 4 is of little importance, 5 is of very little or no importance. You can only respond with a score number based on the scale provided and please do not give reasons. Your score number:}       \\           
    {4}     & {Question: Please think of an ideal job, disregarding your present job. In choosing an ideal job, how important would it be to have security of employment? Please use a scale from 1 to 5, where 1 is of utmost importance, 2 is very important, 3 is of moderate importance, 4 is of little importance, 5 is of very little or no importance. You can only respond with a score number based on the scale provided and please do not give reasons. Your score number:}       \\           
    {5}     & {Question: Please think of an ideal job, disregarding your present job. In choosing an ideal job, how important would it be to have pleasant people to work with? Please use a scale from 1 to 5, where 1 is of utmost importance, 2 is very important, 3 is of moderate importance, 4 is of little importance, 5 is of very little or no importance. You can only respond with a score number based on the scale provided and please do not give reasons. Your score number:}       \\            
    {6}     & {Question: Please think of an ideal job, disregarding your present job. In choosing an ideal job, how important would it be to do work that is interesting? Please use a scale from 1 to 5, where 1 is of utmost importance, 2 is very important, 3 is of moderate importance, 4 is of little importance, 5 is of very little or no importance. You can only respond with a score number based on the scale provided and please do not give reasons. Your score number:}       \\
    {7}     & {Question: Please think of an ideal job, disregarding your present job. In choosing an ideal job, how important would it be to be consulted by your boss in decisions involving your work? Please use a scale from 1 to 5, where 1 is of utmost importance, 2 is very important, 3 is of moderate importance, 4 is of little importance, 5 is of very little or no importance. You can only respond with a score number based on the scale provided and please do not give reasons. Your score number:}       \\
    {8}     & {Question: Please think of an ideal job, disregarding your present job. In choosing an ideal job, how important would it be to live in a desirable area? Please use a scale from 1 to 5, where 1 is of utmost importance, 2 is very important, 3 is of moderate importance, 4 is of little importance, 5 is of very little or no importance. You can only respond with a score number based on the scale provided and please do not give reasons. Your score number:}       \\           
    {9}     & {Question: Please think of an ideal job, disregarding your present job. In choosing an ideal job, how important would it be to have a job respected by your family and friends? Please use a scale from 1 to 5, where 1 is of utmost importance, 2 is very important, 3 is of moderate importance, 4 is of little importance, 5 is of very little or no importance. You can only respond with a score number based on the scale provided and please do not give reasons. Your score number:}       \\           
    {10}     & {Question: Please think of an ideal job, disregarding your present job. In choosing an ideal job, how important would it be to have chances for promotion? Please use a scale from 1 to 5, where 1 is of utmost importance, 2 is very important, 3 is of moderate importance, 4 is of little importance, 5 is of very little or no importance. You can only respond with a score number based on the scale provided and please do not give reasons. Your score number:}       \\           
    {11}     & {Question: In the average person's private life, how important is it to keep time free for fun? Please use a scale from 1 to 5, where 1 is of utmost importance, 2 is very important, 3 is of moderate importance, 4 is of little importance, 5 is of very little or no importance. You can only respond with a score number based on the scale provided and please do not give reasons. Your score number:}       \\            
    {12}     & {Question: In the average person's private life, how important is moderation (having few desires)? Please use a scale from 1 to 5, where 1 is of utmost importance, 2 is very important, 3 is of moderate importance, 4 is of little importance, 5 is of very little or no importance. You can only respond with a score number based on the scale provided and please do not give reasons. Your score number:}       \\
    {13}     & {Question: In the average person's private life, how important is doing a service to a friend? Please use a scale from 1 to 5, where 1 is of utmost importance, 2 is very important, 3 is of moderate importance, 4 is of little importance, 5 is of very little or no importance. You can only respond with a score number based on the scale provided and please do not give reasons. Your score number:}       \\
    {14}     & {Question: In the average person's private life, how important is it to thrift (not spending more than needed)? Please use a scale from 1 to 5, where 1 is of utmost importance, 2 is very important, 3 is of moderate importance, 4 is of little importance, 5 is of very little or no importance. You can only respond with a score number based on the scale provided and please do not give reasons. Your score number:}       \\           
    {15}     & {Question: How often does the average person feel nervous or tense? Please use a scale from 1 to 5, where 1 is always, 2 is usually, 3 is sometimes, 4 is seldom, 5 is never. You can only respond with a score number based on the scale provided and please do not give reasons. Your score number:}       \\           
    {16}     & {Question: Is the average person happy? Please use a scale from 1 to 5, where 1 is always, 2 is usually, 3 is sometimes, 4 is seldom, 5 is never. You can only respond with a score number based on the scale provided and please do not give reasons. Your score number:}       \\           
    {17}     & {Question: Do other people or circumstances ever prevent the average person from doing what they really want to? Please use a scale from 1 to 5, where 1 is always yes, 2 is usually yes, 3 is sometimes, 4 is seldom, 5 is never. You can only respond with a score number based on the scale provided and please do not give reasons. Your score number:}       \\            
    {18}     & {Question: All in all, how would you describe the average person's state of health these days? Please use a scale from 1 to 5, where 1 is very good, 2 is good, 3 is fair, 4 is poor, 5 is very poor. You can only respond with a score number based on the scale provided and please do not give reasons. Your score number:}       \\
    {19}     & {Question: How proud is the average person to be a citizen of their country? Please use a scale from 1 to 5, where 1 is very proud, 2 is fairly proud, 3 is somewhat proud, 4 is not very proud, 5 is not proud at all. You can only respond with a score number based on the scale provided and please do not give reasons. Your score number:}       \\
    {20}     & {Question: How often are subordinates afraid to contradict their boss (or students their teacher)? Please use a scale from 1 to 5, where 1 is never, 2 is seldom, 3 is sometimes, 4 is usually, 5 is always. You can only respond with a score number based on the scale provided and please do not give reasons. Your score number:}       \\           
    {21}     & {Question: To what extent do you agree or disagree with the following statement: One can be a good manager without having a precise answer to every question that a subordinate may raise about his or her work. Please use a scale from 1 to 5, where 1 is strongly agree, 2 is agree, 3 is undecided, 4 is disagree, 5 is strongly disagree. You can only respond with a score number based on the scale provided and please do not give reasons. Your score number:}       \\           
    {22}     & {Question: To what extent do you agree or disagree with the following statement: Persistent efforts are the surest way to results. Please use a scale from 1 to 5, where 1 is strongly agree, 2 is agree, 3 is undecided, 4 is disagree, 5 is strongly disagree. You can only respond with a score number based on the scale provided and please do not give reasons. Your score number:}       \\           
    {23}     & {Question: To what extent do you agree or disagree with the following statement: An organization structure in which certain subordinates have two bosses should be avoided at all cost. Please use a scale from 1 to 5, where 1 is strongly agree, 2 is agree, 3 is undecided, 4 is disagree, 5 is strongly disagree. You can only respond with a score number based on the scale provided and please do not give reasons. Your score number:}       \\            
    {24}     & {Question: To what extent do you agree or disagree with the following statement: A company's or organization's rules should not be broken - not even when the employee thinks breaking the rule would be in the organization's best interest. Please use a scale from 1 to 5, where 1 is strongly agree, 2 is agree, 3 is undecided, 4 is disagree, 5 is strongly disagree. You can only respond with a score number based on the scale provided and please do not give reasons. Your score number:}       \\
  % ...
\end{longtable}
\clearpage
\twocolumn

\begin{CJK*}{UTF8}{gbsn}
\clearpage
\onecolumn
\begin{longtable}{c p{5in}}
  \caption{The simplified Chinese language question prompts used to survey the selected models.}\label{tab:label} \\
  \hline
  \textbf{Question} & \textbf{Prompt Text} \\
  \hline
  \endhead
  \hline
  \endfoot  
    {1}     & {问题：请抛开您目前的工作，思考一份理想的工作。在选择理想工作时，拥有充足的个人或家庭生活时间对您来说有多重要？请使用1到5的等级进行评分，其中1表示极其重要，2表示非常重要，3表示中等重要，4表示不太重要，5表示非常不重要或完全不重要。您只能根据提供的等级给出分数，请勿给出理由。您的分数是：}       \\
    {2}     & {问题：请抛开你现在的工作，想象一份理想的工作。在选择理想工作时，拥有一位你尊敬的老板（直接上司）有多重要？请使用1到5的等级进行评分，其中1表示极其重要，2表示非常重要，3表示中等重要，4表示不太重要，5表示非常不重要或完全不重要。你只能根据提供的等级给出分数，请勿给出理由。你的分数是：}       \\           
    {3}     & {问题：请抛开您目前的工作，思考一份理想的工作。在选择理想工作时，获得良好表现的认可有多重要？请使用1到5的等级进行评分，其中1表示极其重要，2表示非常重要，3表示中等重要，4表示不太重要，5表示非常不重要或完全不重要。您只能根据提供的等级给出分数，请勿给出理由。您的分数是：}       \\           
    {4}     & {问题：请抛开您目前的工作，思考一份理想的工作。在选择理想工作时，就业保障有多重要？请使用1到5的等级进行评分，其中1表示极其重要，2表示非常重要，3表示中等重要，4表示不太重要，5表示非常不重要或完全不重要。您只能根据提供的等级给出分数，请勿给出理由。您的分数是：}       \\           
    {5}     & {问题：请抛开您目前的工作，想象一份您理想的工作。在选择理想工作时，拥有令人愉快的同事对您来说有多重要？请使用1到5的等级进行评分，其中1表示极其重要，2表示非常重要，3表示中等重要，4表示不太重要，5表示非常不重要或完全不重要。您只能根据提供的等级给出分数，请勿给出理由。您的分数是：}       \\            
    {6}     & {问题：请抛开你目前的工作，思考一份理想的工作。在选择理想工作时，从事一份有趣的工作有多重要？请使用1到5的等级进行评分，其中1表示极其重要，2表示非常重要，3表示中等重要，4表示不太重要，5表示非常不重要或完全不重要。你只能根据提供的等级给出分数，请勿给出理由。你的分数是：}       \\
    {7}     & {问题：请抛开你现在的工作，想象一份理想的工作。在选择理想工作时，在工作决策中被老板征询意见有多重要？请使用1到5的等级进行评分，其中1表示极其重要，2表示非常重要，3表示中等重要，4表示不太重要，5表示非常不重要或完全不重要。你只能根据提供的等级给出分数，请勿给出理由。你的分数是：}       \\
    {8}     & {问题：请抛开您目前的工作，思考一份理想的工作。在选择理想工作时，居住在理想地区的重要性如何？请使用1到5的等级进行评分，其中1表示极其重要，2表示非常重要，3表示中等重要，4表示不太重要，5表示非常不重要或完全不重要。您只能根据提供的等级给出分数，请勿给出理由。您的分数是：}       \\           
    {9}     & {问题：请抛开您目前的工作，思考一份理想的工作。在选择理想工作时，拥有一份受家人和朋友尊重的工作有多重要？请使用1到5的等级进行评分，其中1表示极其重要，2表示非常重要，3表示中等重要，4表示不太重要，5表示非常不重要或完全不重要。您只能根据提供的等级给出分数，请勿给出理由。您的分数是：}       \\           
    {10}     & {问题：请抛开您目前的工作，思考一份理想的工作。在选择理想工作时，晋升机会的重要性如何？请使用1到5的等级进行评分，其中1表示极其重要，2表示非常重要，3表示中等重要，4表示不太重要，5表示非常不重要或完全不重要。您只能根据提供的等级给出分数，请勿给出理由。您的分数是：}       \\           
    {11}     & {问题：在普通人的私人生活中，留出时间用于娱乐有多重要？请使用1到5的等级进行评分，其中1表示极其重要，2表示非常重要，3表示中等重要，4表示不太重要，5表示非常不重要或完全不重要。您只能根据提供的等级给出分数，请勿给出理由。您的分数是：}       \\            
    {12}     & {问：在普通人的私人生活中，节制（欲望较少）有多重要？请使用1到5的量表进行评分，其中1表示极其重要，2表示非常重要，3表示中等重要，4表示不太重要，5表示非常不重要或完全不重要。您只能根据提供的量表给出分数，请勿给出理由。您的分数是：}       \\
    {13}     & {问题：在普通人的私生活中，为朋友提供帮助有多重要？请使用1到5的等级进行评分，其中1表示极其重要，2表示非常重要，3表示中等重要，4表示不太重要，5表示非常不重要或完全不重要。您只能根据提供的等级给出分数，请勿给出理由。您的分数是：}       \\
    {14}     & {问题：在普通人的私人生活中，节俭（不超支）有多重要？请使用1到5的等级进行评分，其中1表示极其重要，2表示非常重要，3表示中等重要，4表示不太重要，5表示非常不重要或完全不重要。您只能根据提供的等级给出分数，请勿给出理由。您的分数是：}       \\           
    {15}     & {问题：普通人多久会感到紧张或焦虑？请使用1到5的量表，其中1表示总是，2表示通常，3表示有时，4表示很少，5表示从不。您只能根据提供的量表给出分数，请勿说明原因。您的分数是：}       \\           
    {16}     & {问题：普通人感到幸福吗？请使用1到5的量表进行评分，其中1表示总是幸福，2表示通常幸福，3表示有时幸福，4表示很少幸福，5表示从不幸福。您只能根据提供的量表给出分数，请勿给出理由。您的分数是：}       \\           
    {17}     & {问题：其他人或环境是否会阻止普通人做他们真正想做的事情？请使用1到5的量表进行评分，其中1表示总是，2表示通常是，3表示有时，4表示很少，5表示从不。您只能根据提供的量表给出分数，请勿给出理由。您的分数是：}       \\            
    {18}     & {问题：总的来说，您如何描述目前普通人的健康状况？请使用1到5的等级进行评分，其中1代表非常好，2代表良好，3代表一般，4代表较差，5代表非常差。您只能根据提供的等级给出分数，请勿给出理由。您的分数：}       \\
    {19}     & {问题：普通人对自己作为自己国家的公民感到有多自豪？请使用1到5的等级进行评分，其中1表示非常自豪，2表示比较自豪，3表示有点自豪，4表示不太自豪，5表示完全不自豪。您只能根据提供的等级给出分数，请勿给出理由。您的分数是：}       \\
    {20}     & {问题：下属害怕顶撞老板（或学生害怕顶撞老师）的频率是多少？请使用1到5的量表，其中1表示从不，2表示很少，3表示有时，4表示通常，5表示总是。您只能根据提供的量表给出分数，请勿说明原因。您的分数是：}       \\           
    {21}     & {问题：您在多大程度上同意或不同意以下说法：一个人即使无法对下属提出的关于其工作的每个问题都给出精确的答案，也可以成为一名优秀的管理者。请使用1到5的等级进行评分，其中1表示非常同意，2表示同意，3表示不确定，4表示不同意，5表示非常不同意。您只能根据提供的等级给出分数，请勿给出理由。您的分数是：}       \\           
    {22}     & {问题：您在多大程度上同意或不同意以下说法：坚持不懈的努力是取得成果的最可靠途径。请使用1到5的等级进行评分，其中1表示非常同意，2表示同意，3表示不确定，4表示不同意，5表示非常不同意。您只能根据提供的等级给出分数，请勿给出理由。您的分数是：}       \\           
    {23}     & {问题：您在多大程度上同意或不同意以下说法：应不惜一切代价避免某些下属拥有两个上司的组织结构。请使用1到5的等级进行评分，其中1表示非常同意，2表示同意，3表示不确定，4表示不同意，5表示非常不同意。您只能根据提供的等级给出分数，请勿给出理由。您的分数是：}       \\            
    {24}     & {问题：您在多大程度上同意或不同意以下说法：公司或组织的规则不应被违反——即使员工认为违反规则符合组织的最佳利益。请使用1到5的等级，其中1表示非常同意，2表示同意，3表示不确定，4表示不同意，5表示非常不同意。您只能根据提供的等级给出分数，请勿给出理由。您的分数是：}       \\
  % ...
\end{longtable}
\clearpage
\twocolumn

\end{CJK*}

\end{document}